\documentclass[letterpaper]{article} % DO NOT CHANGE THIS
\usepackage{aaai25}  % DO NOT CHANGE THIS
\usepackage{times}  % DO NOT CHANGE THIS
\usepackage{helvet}  % DO NOT CHANGE THIS
\usepackage{courier}  % DO NOT CHANGE THIS
\usepackage[hyphens]{url}  % DO NOT CHANGE THIS
\usepackage{graphicx} % DO NOT CHANGE THIS
\usepackage{natbib}  % DO NOT CHANGE THIS AND DO NOT ADD ANY OPTIONS TO IT
\usepackage{caption} % DO NOT CHANGE THIS AND DO NOT ADD ANY OPTIONS TO IT
\usepackage{algorithm}
\usepackage{algorithmicx}
\usepackage{algpseudocode}
\usepackage{textgreek}
\usepackage[table,xcdraw]{xcolor}

\usepackage{newfloat}
\usepackage{listings}
\DeclareCaptionStyle{ruled}{labelfont=normalfont,labelsep=colon,strut=off} % DO NOT CHANGE THIS
\floatstyle{ruled}
\newfloat{listing}{tb}{lst}{}
\floatname{listing}{Listing}
\usepackage{amsmath,comment,amsthm,amssymb}
\usepackage{multirow}

\newtheorem{lemma}{Lemma}
\newtheorem{theorem}{Theorem}
\newtheorem{definition}{Definition}

\usepackage{kotex}

\usepackage{xcolor}
\definecolor{emerald}{RGB}{26,121,42}
\definecolor{topaz}{RGB}{236,185,57}
\definecolor{sapphire}{RGB}{14,26,164}

\usepackage[colorinlistoftodos,prependcaption,obeyFinal]{todonotes}
\usepackage{multirow}
\usepackage{caption}
\presetkeys{todonotes}{inline, color=topaz!40, bordercolor=topaz}{}

\nocopyright % -- Your paper will not be published if you use this command
\title{Towards Efficient Formal Verification of Spiking Neural Network}
\author{
    Baekryun Seong\textsuperscript{\rm 1}, Jieung Kim\textsuperscript{\rm 2}, and Sang-Ki Ko\textsuperscript{\rm 1}
}
\affiliations{
    \textsuperscript{\rm 1}Department of Artificial Intelligence,
    University of Seoul, Seoul, South Korea\\
    \{sizzflair97,sangkiko\}uos.ac.kr\\
    \textsuperscript{\rm 2}Department of Computer Science,
    Yonsei University, Seoul, South Korea\\
    jieungkim@yonsei.ac.kr
}

\usepackage{bibentry}
\begin{document}

\maketitle

\begin{abstract}
Recently, AI research has primarily focused on large language models (LLMs), and increasing accuracy often involves scaling up and consuming more power. The power consumption of AI has become a significant societal issue; in this context, spiking neural networks (SNNs) offer a promising solution. SNNs operate event-driven, like the human brain, and compress information temporally. These characteristics allow SNNs to significantly reduce power consumption compared to perceptron-based artificial neural networks (ANNs), highlighting them as a next-generation neural network technology.
However, societal concerns regarding AI go beyond power consumption, with the reliability of AI models being a global issue. For instance, adversarial attacks on AI models are a well-studied problem in the context of traditional neural networks. Despite their importance, the stability and property verification of SNNs remains in the early stages of research. Most SNN verification methods are time-consuming and barely scalable, making practical applications challenging.
In this paper, we introduce temporal encoding to achieve practical performance in verifying the adversarial robustness of SNNs. We conduct a theoretical analysis of this approach and demonstrate its success in verifying SNNs at previously unmanageable scales. Our contribution advances SNN verification to a practical level, facilitating the safer application of SNNs.
\end{abstract}

% Uncomment the following to link to your code, datasets, an extended version or similar.
%
% \begin{links}
%     \link{Code}{https://aaai.org/example/code}
%     \link{Datasets}{https://aaai.org/example/datasets}
%     \link{Extended version}{https://aaai.org/example/extended-version}
% \end{links}

\section{Introduction}

Currently, AI research is primarily focused on large language models (LLMs) since the advent of powerful models such as OpenAI's ChatGPT~\cite{brown2020language}. It is well known that the performance of AI models generally improves as the number of parameters increases~\cite{montufar2014number,goodfellow2013multi}. However, this also leads to a significant increase in power consumption~\cite{desislavov2023trends,samsi2023words}, which leads to environmental side effects such as carbon emissions and climate change.

Spiking neural networks (SNNs) are neural networks that mimic the human brain, and computing SNNs by mimicking neurons in electronic circuits consumes less power than typical perceptron-based artificial neural networks (ANNs)~\cite{kim2020spiking}. % However, research on SNNs is still in its infancy, and the black-box nature of neural networks has slowed their adoption.
Many researchers claim that SNNs are the future of AI research and could be an effective alternative to address the excessive power consumption problem caused by current ANNs~\cite{eshraghian2023training,tavanaei2019deep}. However, studies supporting these claims are still in their early stages.

On the other hand, many people suggested the potential risk of adversarial attacks on neural networks by introducing a carefully designed perturbation to the input data, which causes severe problems with neural networks' security and safety~\cite{akhtar2018threat,moosavi2017universal,szegedy2013intriguing}. 
Since neural networks are essentially function approximators~\cite{hornik1989multilayer}, it is important to verify the approximation ratio, which is the range of error in their output, to estimate the robustness of the networks to be deployed in safety-critical applications. To date, there have been a lot of attempts to formally verify ANNs~\cite{albarghouthi2021introduction,katz2022reluplex,tjeng2017evaluating,bunel2020branch}.

However, it is still an early stage for SNNs, and several researchers have only suggested the plausibility of formal verification of SNNs~\cite{de2020spiking,BanerjeeGBM23,de2017formal}.
Recently, there has been a promising study that aims to verify the local adversarial robustness of SNNs using satisfiability modulo theories (SMT) solvers~\cite{BanerjeeGBM23}. 
They have shown that it is theoretically possible to verify the robustness of SNNs, but it is notoriously difficult to apply in practice due to its high computational cost, even at small scales.
It is not difficult to speculate that the rigorous formal verification of SNNs is generally more difficult than that of traditional ANNs because of the temporal and discrete nature of the network. 
%While the input values and activations are real numbers in ANNs..
%\sk{이 부분 다시 써야할 듯..}, and the propagation of activations ends at a single moment. Also, none of the perceptrons store previous activations. 
% However, the input of an SNN takes many steps, the input is encoded in a time scale, and the neurons of an SNN remember the activations propagated at previous times by converting them into action potentials. 
For instance, the state space of SNNs involves one more dimension, the temporal dimension, as the input is described as a spike train along the time axis rather than an instantaneous one. In addition, the activation of SNNs is a discrete function of the potential, which is not differentiable, unlike ANNs. These two properties make rigorous verification and analytic interpretation more difficult for SNNs than for ANNs.

In this paper, we present an efficient verification algorithm for SNNs, and the following are our key contributions:
\begin{enumerate}
    \item Formulate spiking neural network with temporal encoding as SMT solver constraints.
    \item Theoretically analyze the impact of temporal encoding on the speed of formal verification.
    \item Empirically measure and compare the verification times and propose an algorithm to overcome the limitations of the SMT solver.
\end{enumerate}
% Figure~\ref{fig:overview} shows the graphical description of our key contributions.
%ompares the encodings and shows the structure of the temporal-encoded SNN. Our contributions will help train SNNs to be more robust, and by speeding up formal verification, they will help SNNs be applied in the real world.
% \sk{전반적으로 reference 양이 너무 적다는 느낌을 받습니다. 보통 20~40개 정도는 있는 것 같아요. 관련 연구들을 많이 찾아 포함하는 게 나아보입니다.}

\section{Related Work}
% \br{ASE Review - Although Section 2 presents related literature with sufficient detail, the similarities and differences between the approach proposed by the authors and prior verification approaches need to be explicitly outlined. This comparison is necessary to achieve a clearer understanding of the foundational research from which the proposed approach originates, delineating the specific innovations and unique contributions introduced by the authors.}

\subsection{Formal Verification of Neural Networks}
% \jk{Related work 이 contraint-based solving 쪽에 집중 되어 있는 느낌인데, 혹시 contraint based solving이 SNN verification에 더 비슷하기 때문일까요? contraint-based solving과 abstraction-based solving으로 categorize 가능한데, 이 중 constraint-based solving에 집중한다는 것을 언급을 해 주는 것이 좋을까요?}
% \br{Verification categories에 대한 서술 추가했습니다.}
Formal verification of neural networks is an active area of research, mainly related to robustness~\cite{albarghouthi2021introduction}. There are two main ways to verify the properties of neural networks, abstraction-based verification and constraint-based verification. Methods based on constraint-based verification approximate the neural network using abstract interpretation theory.  Katz et al.~\cite{katz2022reluplex} have extended the simplex method to perform verification for neural networks with ReLU activation functions, which are frequently employed in deep neural networks. Later, Katz et al.~\cite{KatzHIJLLSTWZDK19} introduced the Marabou framework by extending Reluplex by considering arbitrary piecewise-linear activation functions.
Tjeng et al.~\cite{tjeng2017evaluating} have utilized mixed-integer linear programming (MILP) to represent ReLU and achieved better robustness verification performance than Reluplex. Bunel et al.~\cite{bunel2020branch} have utilized a branch-and-bound algorithm to perform formal verification. Most of abstraction-based verification methods for ANNs are based on piecewise linear activation in real space, which is hard to be applied to the discrete spikes of SNNs. On the other hand, constraint-based methods use SMT solver, which can also applied in discrete activations. Pulina and Tacchella used an SMT solver to verify the properties of ANN~\cite{pulina2012challenging}. Amir et al.~\cite{Amir0BK21} applied an SMT solver to verify the properties of a binarized neural network, extending ReLUplex. Guo et al.~\cite{GuoZZKZ23} proposed an efficient SMT-based verification method for occlusion robustness.

\subsection{Mathematical Modeling of SNNs}
Due to its event-driven and time-varying nature, the mathematical models for SNNs differ substantially from those for ANNs. De Maria et al.~\cite{de2017formal,de2020spiking} have proposed a formalization of SNNs based on timed automata~\cite{alur1994theory,waez2013survey} networks. As each neuron of SNN can be modeled by a timed automaton, they have constructed a network that consists of a set of timed automata running in parallel and sharing channels according to the structure of the SNN. The authors have validated the constructed model against several properties of SNNs via temporal logic formulae.

More recently, SMT-based encoding of SNN for adversarial robustness verification has been introduced~\cite{BanerjeeGBM23}. The authors have adapted the quantifier-free linear real arithmetic constraints to encode the overall mathematical operations in SNNs. They have further encoded the adversarial robustness conditions based on potential perturbations on input as logical constraints and demonstrated that SMT solvers can be successfully used for modeling and verifying the robustness of SNNs. However, the experiments are conducted on relatively small SNNs with a restricted benchmark due to the limited scalability of the proposed approach.

\subsection{Information Encoding in SNNs}
%SNN에서 입력 데이터를 변환하는 방식은 크게 rate encoding과 temporal encoding 등이 있다. Rate 인코딩은 일반적으로 SNN에서 많이 쓰이는 Input data encoding 방법으로, Input의 크기를 spike할 확률로 해석하여 이 확률에 따라 random한 스파이크를 생성한다. 정보의 크기가 spike 수에 비례한다는 간단한 직관을 따르는 이 방법은 구현이 간편하나, step별로 계속해서 spike가 생성되고, SNN을 구현하는 neuromorphic chip은 spike가 증가할수록 많은 전력을 소모하기 때문에 더 많은 전력을 소모한다. Temporal encoding은 일반적으로 각 Input neuron이 한 번만 encoding하도록 하며, Input의 크기가 클수록 더 빨리 스파이크하도록 변환한다. Temporal encoding은 훨씩 적은 전력을 소모하지만, ANN의 경사 하강을 그대로 사용하기 어렵기 때문에 일반적으로 훈련이 어렵다.
Three main ways to convert input data in SNNs are rate encoding, temporal encoding, and delta modulation~\cite{eshraghian2023training}. Rate encoding is a commonly used method in SNNs that interprets the size of the input as the probability of spiking and generates random spikes based on this probability~\cite{adrian1926impulses}. This method, which follows the simple intuition that the size of the information is proportional to the number of spikes, is simple to implement because the gradient descent of the ANN can be applied through a surrogate derivative~\cite{guerguiev2017towards}. However, it consumes more power because the number of spikes is large, and neuromorphic chips implementing SNNs consume more power as the number of spikes increases. Temporal encoding generally forces each input neuron to encode only once, and the larger the input, the faster it spikes~\cite{johansson2004first,gollisch2008rapid}. It consumes significantly less power but is generally harder to train because it is harder to use the gradient descent of an ANN as is. Delta modulation encodes data to spike when it increases over time. It requires sequential data, so it is not used in most cases~\cite{petro2019selection}.
Mostafa~\cite{mostafa2017supervised} introduced the gradient descent learning algorithm on temporal-encoded SNN, using the property that the input-output latency relation in temporal-encoded SNN is differentiable almost everywhere. Kheradpisheh et al.~\cite{S4NN} improved Mostafa's algorithm~\cite{mostafa2017supervised} and published an implementation in a public online repository. Yamamoto et al.~\cite{yamamoto2022timing} proposed temporal encoding without single spike restriction.

\subsection{Adversarial Robustness in SNNs}
The adversarial robustness of neural networks is important in practical use, so various research studies have been conducted to increase the adversarial robustness of SNNs, like ANNs. Zhou~\cite{zhou2021temporal} has suggested the method to train robust temporal encoded SNN. Liang et al. designed S-IBP and S-CROWN for robust training of SNN.~\cite{liang2022toward} Özdenizci et al. presented an algorithm that converts ANNs to SNNs and fine-tunes it to strengthen robustness. ~\cite{ozdenizci2023adversarially} Ding et al. showed a stochastic gating spiking neural model, which is biologically plausible, to increase robustness.~\cite{ding2024enhancing} Chen et al. proposed a way to train robust SNN using the inspiration from the visual masking effect and filtering theory.~\cite{chen2024defendingspikingneuralnetworks}

\section{Preliminaries}
\subsection{Spiking Neural Networks}
\paragraph{Variable definitions}
We need to define variables to describe SNN. $T$ is the number of simulated time steps, $L$ is the total number of layers where the first layer is the input layer, and $N_l$ is the number of neurons in layer $l$. For simplicity, we write the size of the input as $N_0$. The spikes of SNN is denoted by $\mathbf{x}_{l}^{(t)} = \left(x_0^{(t)}, x_1^{(t)}, \ldots, x_{N_l -1}^{(t)}\right) \in \{0,1\}^{N_l}$. Lastly, $\tau$, which is set to $1$ here, indicates the {\em synaptic delay} that refers to the time it takes for a spike to be transmitted from a pre-synaptic neuron to the postsynaptic neuron.

\begin{figure}[t!]
\centering
\includegraphics[width=.47\textwidth, clip, trim={0.4cm 0.6cm 0.8cm 0.4cm}]{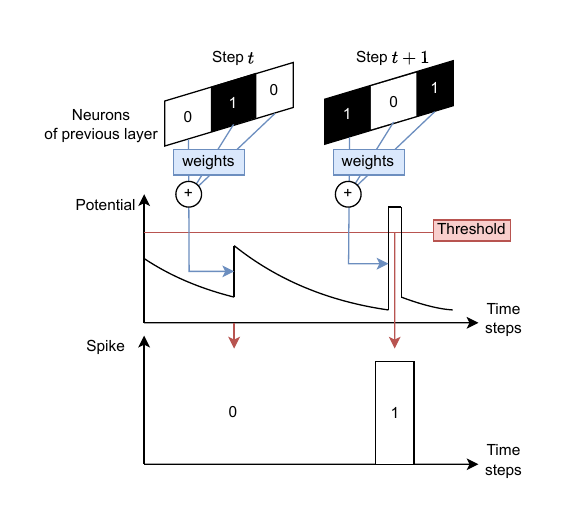}
\caption{The mechanism of SNN. The Spike train of the previous layer, which consists of 0 and 1, is multiplied with weights and integrated together. The integrated value is added to potential, which decays with time.}
\label{fig:snn-desc}
\end{figure}

\paragraph{Spiking neurons}
{\em Leaky Integrate-and-Fire} (LIF) neurons and {\em Integrate-and-Fire} (IF) neurons are artificial neuron models that integrate binary presynaptic input spikes over time and fire to postsynaptic neurons if potential exceeds the threshold. Figure~\ref{fig:snn-desc} shows the overview of SNN. In general, a spike firing is represented as $1$, and $0$ otherwise. On discrete time steps, we can define LIF neurons in the form of multivariate functions that map the state of the past to the future. First, the LIF neuron integrates the input current and some of its past potential $p^{(t-1)}$.
\begin{align*}
    &\text{Integrate}(\mathbf{x}, \mathbf{w}, p^{(t-1)};\gamma)=
        \langle \mathbf{x},\mathbf{w} \rangle +\gamma p^{(t-1)}
\end{align*}
Then, the neuron determines whether to generate a spike according to the integrated current and updates its potential. Note that the spike timing of the next layer depends on the synaptic delay $\tau$, but the potential does not.
\begin{align*}
    x_{\text{out}}^{(t)}&=\begin{cases}
        1\quad\text{if\ \ }\text{Integrate}\left(\mathbf{x}_{\text{in}}^{(t-\tau)}, p^{(t-\tau-1)};\mathbf{w}_{\text{in}},\gamma\right)\ge\theta\\
        0\quad\text{otherwise.}
        \end{cases}\\
    p^{(t)} &= \text{Integrate}\left(\mathbf{x}_{\text{in}}^{(t)}, \mathbf{w}_{\text{in}}, p^{(t-1)};\gamma\right)-\theta x_{\text{out}}^{(t)}
\end{align*}
By substituting $\gamma$ to 1, they become the definition of IF neurons. Our method uses IF neurons due to the ease of training.
A {\em spiking neural network} (SNN) is a neural network that employs biologically plausible neuron models such as LIF and IF neurons. In general, SNNs are simulated over many time steps to generate outputs.

\begin{figure}[t!]
\centering
\includegraphics[width=.47\textwidth, clip, trim={0.8cm 0.6cm 1cm 0.8cm}]{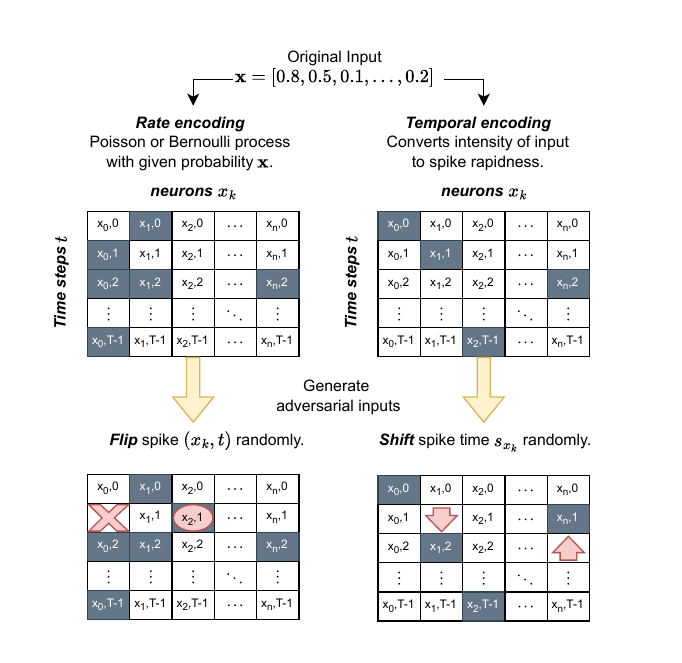}
\caption{Two ways to encode inputs in SNNs.}
\label{fig:enc-ex}
\end{figure}
\paragraph{Spike encoding}
Due to the nature of SNNs described above, inputs targeting any SNN must be encoded in temporal and binary form, and we must interpret the output spike train for prediction. We utilize two ways to encode inputs: rate and temporal (latency) encoding.

In SNNs, rate encoding generates a spike train based on the Poisson or Bernoulli random process, whereby inputs are interpreted as probabilities. If an input is not between 0 and 1, it must be scaled to fit within this range. On the other hand, temporal encoding makes a spike train by interpreting the intensity of inputs to input quickness. The larger the inputs, the faster the spikes. The upper side of Figure~\ref{fig:enc-ex} shows the difference between two input encodings.

Given that the meanings of time steps vary according to input encoding, the methods for making predictions based on the output must differ. In rate encoding, we count each neuron's spikes and choose the neuron with the most spikes. In contrast, in temporal encoding, we utilize the {\em time-to-first-spike} (TTFS) method, which selects the fastest spiking neuron. (Note that neurons in the hidden and output layer can spike multiple times.) Taking TTFS into account, we do not need whole spike train $\mathbf{x}_{l}^{(t)}$. Instead, we can use the {\em spike times} $\mathbf{s}_{l}\in\mathbb{N}^{N_l}$.

\subsection{Adversarial Attack}
In ANNs, adversarial attacks are generated by adding a real-valued perturbation to the original input. However, we cannot use this approach to SNNs because the spike trains do not consist of a real number but binary. Moreover, considering that the inputs of rate are far from the inputs of temporal encodings, the way to generate adversarial attacks must be different. 

\paragraph{Spike train perturbation}
In rate encoding, previous work ~\cite{BanerjeeGBM23} proposed the definition of perturbation based on L1 distance. The elements in the spike train are randomly flipped, with the number of flips being less than or equal to the value of $\Delta$.

\begin{figure}[t!]
\centering
\includegraphics[width=.47\textwidth, clip, trim={0.4cm 0.2cm 0 0.8cm}]{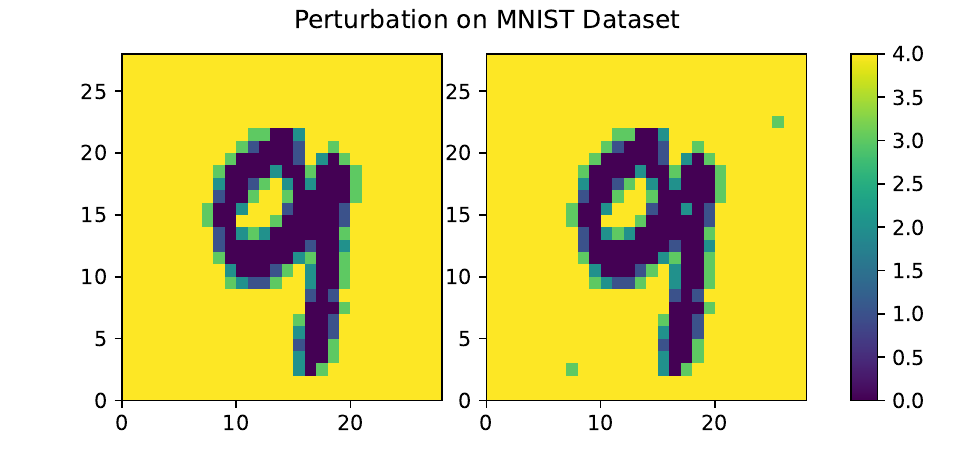}
\caption{MNIST Examples of the original input image (left) and the image with temporal perturbation (right) where $\Delta=4$.
}
\label{fig:pert-ex}
\end{figure}

However, it is not possible to employ this method in temporal encoding, as it is essential that all neurons in the input layer fire only once; a random flip could corrupt this property. Instead, we randomly shift the spike times, and the sum of the distances between original and perturbed spike times is less than or equal to $\Delta$. The lower side of Figure~\ref{fig:enc-ex} shows examples of adversarial attacks. Figure~\ref{fig:pert-ex} shows an example of an input spike times and adversarial attack in temporal encoding.

\paragraph{Perturbation space}
For a theoretical analogy, we need a measure to compare two different encodings. $\Delta$-perturbation space is the size of the set of possible perturbed inputs. We have a different definition of perturbation set for the two encoding cases because the spike can not be flipped in temporal encoding.
\begin{definition}[$\Delta$-Perturbation Space]\label{def:delta-perturbation-rate}
For a non-negative integer~$\Delta$ the $\Delta$-perturbation set of a rate-encoded input spike train $X$ is,
\begin{align*}
    \varepsilon(X, \Delta) \equiv \left\{\tilde{X}\ \middle|\ \sum_{n=0}^{N_0-1}\left\|\tilde{\mathbf{x}}_{n}-\mathbf{x}_{n}\right\|_1\le \Delta\right\}.
\end{align*}
Furthermore, the $\Delta$-perturbation set of an input spike times $\mathbf{s}$ is,
\begin{align*}
    \varepsilon(\mathbf{s}, \Delta) \equiv \left\{\tilde{\mathbf{s}}\ \vline\ \sum_{n=0}^{N_0-1}|\tilde{s}_{n}-s_{n}|\le \Delta\right\}.
\end{align*}
Then, the $\Delta$-perturbation space of the input spike train or spike times is defined as the size of the $\Delta$-perturbation set of it.
\end{definition}
% \paragraph{Perturbation space}
% For a theoretical analogy, we need a measure to compare two different encodings. $\Delta$-perturbation space is the size of the set of possible perturbed inputs. The formal definition of rate encoding is below.
% \begin{definition}[$\Delta$-Perturbation Space of Rate Encoding]\label{def:delta-perturbation-rate}
% For a non-negative integer~$\Delta$, the $\Delta$-perturbation set of an input spike train $X$ is,
% \begin{align*}
%     \varepsilon(X, \Delta) \equiv \left\{\tilde{X}\ \middle|\ \sum_{n=0}^{N_0-1}\left\|\tilde{\mathbf{x}}_{n}-\mathbf{x}_{n}\right\|_1\le \Delta\right\}.
% \end{align*}

% Then, the $\Delta$-perturbation space of the input spike train is defined as the size of the $\Delta$-perturbation set of the input spike train.
% \end{definition}
% Instead of using this measure, we define the new one for temporal encoding because the spike can not be flipped in temporal encoding.
% \begin{definition}[$\Delta$-Perturbation Space of Temporal Encoding]\label{def:delta-perturbation-temp}
% For a non-negative integer~$\Delta$, the $\Delta$-perturbation set of an input spike times $\mathbf{s}$ is,
% \begin{align*}
%     \varepsilon(\mathbf{s}, \Delta) \equiv \left\{\tilde{\mathbf{s}}\ \vline\ \sum_{n=0}^{N_0-1}|\tilde{s}_{n}-s_{n}|\le \Delta\right\}.
% \end{align*}
% Then, the $\Delta$-perturbation space of the input spike times is defined as the size of the $\Delta$-perturbation set of the input spike times.
% \end{definition}
Considering that spike times have a similar meaning to ANNs' intensity, the L1 distance between the perturbed spike times and the original is very similar to formal verification in ANNs. Assuming that the spike ratio in rate encoding is an estimate of ground truth probability, the meanings of L1 distance from two different encodings are almost the same. However, the space complexity of the two encodings is very different, as shown in the theoretical analysis section, and it is one of the main contributions of our paper.

\section{Methodology}
First, we introduce our SMT encoding for temporal encoded SNN. We describe mathematical behavior and adversarial robustness conditions in quantifier-free linear real arithmetic formulas for the sound and complete verification of SNNs. Second, we theoretically compare the proposed encoding and the previous SMT encoding~\cite{BanerjeeGBM23} for SNNs with rate encoding in terms of combinatorial complexity of adversarial perturbations on input.

\subsection{SMT Encoding of Temporal Encoded SNN}
We encode expressions to describe SNN, which consists of IF neurons. In the following paragraphs, we denote the $n$th neuron in the $l$th layer by the pair $(n,l)$.
% Let $T$ be the number of simulated time steps, $L$ be the total number of layers where the first layer is the input layer, and $N_l$ be the number of neurons in layer $l$. For simplicity, we write the size of the input as $N_0$. Hence, the input of SNN is denoted by $\mathbf{x} = (x_0, x_1, \ldots, x_{N_0 -1})$, where $x_i$ for $0 \le i \le N_0-1$ belongs to $\mathbb{Z}^T$. Then, we need to define variables to describe SNN. In the following paragraphs, we denote the $n$th neuron in the $l$th layer by the pair $(n,l)$.
% Lastly, $\tau$, which is set to $1$ here, indicates the {\em synaptic delay} that refers to the time it takes for a spike to be transmitted from a pre-synaptic neuron to the postsynaptic neuron.
{\em Spike time} $s_{l,n}$ of the neuron $(n,l)$ is described by a non-negative integer with the following constraint:
\begin{align*}
    &\xi_1\equiv \bigwedge_{l=1}^{L}
                   \bigwedge_{n=0}^{N_l-1}
                     \left[
                       s_{l,n} \ge \tau l \wedge s_{l,n} \le T-1 
                       \right].
\end{align*}
The condition $\xi_1$ asserts that spike time must be in the range $[\tau l, T-1]$ since we assume that there are synaptic delays in the spike time at each layer. Simply speaking, a neuron $(n, l)$ for all $0 \le n \le N_l-1$ cannot spike before $\tau l$ as it requires at least $l$ synaptic delays.

{\em Potential} $p_{l,t,n}$ of the neuron $(n,l)$ at time step $t$ is a real number satisfying the following conditions:
\begin{align*}
    &\xi_2\equiv \bigwedge_{l=1}^{L}
                   \bigwedge_{n=0}^{N_l-1}
                     \left[
                       p_{l,0,n}=0
                     \right]\\
    &\xi_3\equiv \bigwedge_{l=1}^{L}
                   \bigwedge_{n=0}^{N_l-1}
                     \bigwedge_{t=1}^{T-1}
                       \left[
                         p_{l,t,n} =
                           \sum_{m=0}^{N_{l-1}-1}
                             w_{l-1,m,n}\cdot\mathbf{1}(s_{l-1, m}\le t)
                       \right]
\end{align*}
where $\mathbf{1}$ is an indicator function and $w_{l,m,n}$ is a learnable weight of the neuron $(n, l)$ to the neuron $(n, l+1)$. $\xi_2$ and $\xi_3$ assert that neurons integrate the sum of weighted spikes from the previous layer.
%\jk{Currents가 어떠한 의미 인가요? 현재 뉴런을 이야기 하는걸까요?} \br{수정했습니다!}
\begin{align*}
    &\xi_4\equiv \bigwedge_{l=1}^{L}
                   \bigwedge_{n=0}^{N_l-1}
                     \bigwedge_{t=1}^{T-1}
                       \left[
                         a_{l,t,n} =
                         \bigvee_{t'=0}^{t-1} p_{l,t',n}\ge\theta
                       \right]\\
    &\xi_5\equiv \bigwedge_{l=1}^{L}
                   \bigwedge_{n=0}^{N_l-1}
                     \bigwedge_{t=\tau l}^{T-2}
                       \left[
                         \left(
                           \neg a_{l,t-\tau,n}
                           \wedge
                           p_{l,t-\tau,n}\ge\theta
                           \right)
                         =
                         \left(
                           s_{l,n}=t
                           \right)
                         \right]\\
    &\xi_6\equiv \bigwedge_{l=1}^{L}
                   \bigwedge_{n=0}^{N_l-1}
                     \left[
                       \neg a_{l,T-1-\tau,n}
                       =
                       \left(
                         s_{l,n}=T-1
                         \right)
                       \right]
\end{align*}

The condition $\xi_4$ utilizes a flag variable~$a_{l,t,n}$ to indicate whether the neuron $(n,l)$ has ever spiked before time step $t-1$ and $\xi_5$ states that if $(n,l)$ has not spiked before time step $t-\tau-1$ and its potential reaches the threshold at $t-\tau$, it spikes at $t$. The condition $\xi_6$ enforces the neuron $(n, l)$ to spike at the last time step $T-1$ if $(n,l)$ has never spiked before.

{\em Temporal perturbation} about an input $\mathbf{x}$ is expressed as:
\begin{align*}
    &\xi_7\equiv
      \sum_{n=0}^{N_0-1}
        |s_{0,n}-x_{n}|
      \le \Delta.
\end{align*}

The condition $\xi_7$ asserts a constraint that L1 distance between $\mathbf{s}_{0}$ and $\mathbf{x}$ is not greater than $\Delta$.
SNN makes the {\em prediction} $\hat{y}$ for an input $\mathbf{x}$ by choosing the neuron with the fastest input response. 

{\em Local robustness} for the input $\mathbf{x}$ and prediction $\hat{y}$ is encoded as:
\begin{align*}
    &\xi_8\equiv
      \bigwedge_{n=0}^{N_L-1}
        \left[n\ne\hat{y}\implies s_{L,n}>s_{L,\hat{y}}\right].
\end{align*}

We can verify whether or not a given SNN is robust by checking if $\xi_1\wedge\xi_2\wedge\cdots\wedge\xi_7\wedge\neg\xi_8$ is satisfiable using the SMT solver. Namely, the SNN model is not robust if the condition is satisfiable as it implies the existence of an adversarial counterexample.

\subsection{Theoretical Analysis}
Here, we provide a theoretical explanation for why our SMT encoding is more efficient regarding the adversarial robustness verification performance than the one formulated in the previous work~\cite{BanerjeeGBM23}.
The main intuition is to compare the {\em perturbation space}, which is the number of possible adversarial $\Delta$-perturbations on input, of the rate and temporal encoding.
Now, we are ready to provide the theoretical analysis of the exponential complexity advantage by adopting temporal encoding instead of rate encoding. %Note that we omit the full proofs due to space limits.
\begin{lemma}
\label{lemma:pert-space-ratio}
     Let $\Delta=\alpha TN$, where $0\le \alpha\le1$. The perturbation space ratio of rate encoding to temporal encoding is $O((T^{\alpha T}/(1+2\alpha T))^{N})$.
\end{lemma}
\begin{proof}
Let us assume the input layers have the number of neurons $N$.
The perturbation space of rate encoding can be calculated by counting the ways to flip less than or equal to $\Delta$ spikes as follows:
\begin{align*}
    \sum_{d=1}^\Delta 
        \begin{pmatrix}
            NT\\d
        \end{pmatrix}
        % =&\ NT + \cdots + \frac{NT\cdots(NT-\Delta+1)}{\Delta!}\\
    =&\ O\left(\frac{(NT)^\Delta}{\Delta!}\right)
\end{align*}

We can also calculate the perturbation space of temporal encoding. We define {\em delta partitions} $D(\Delta)$ and count the number of the delta partitions sequences as we can distribute $\Delta$ to $N$ input neurons:
\begin{align*}
    D(\Delta) =& \left\{\{\Delta_{n}\}_{n=0}^{N-1}\in\mathbb{N}^N\middle|\sum_{n=0}^{N-1}\Delta_{n}=\Delta\right\}\\
    \left|D(\Delta)\right|=&\begin{pmatrix}
                                N+\Delta-1\\\Delta
                                \end{pmatrix} =\ O\left(\frac{N^\Delta}{\Delta!}\right).\\
                       %   =&\ \frac{(N+\Delta-1)\cdots N}{\Delta!}\\
                        %   =&\ O\left(\frac{N^\Delta}{\Delta!}\right).
\end{align*}
We can get {\em delta sequences} $S(\cdot)$ along to input neurons, for each sequence in a delta partition $\{\Delta_{n}\}_{n=0}^{N-1}$. We can estimate the upper bound using the Lagrangian multiplier method with logarithm:
\begin{align*}
    S\left(\left\{\Delta_{n}\right\}_{n=0}^{N-1}\right)=&
        \left\{\left\{\delta_{n}\right\}_{n=0}^{N-1}\in\mathbb{Z}^N \middle|\ 
        \forall_n\left|\delta_n\right|\le\Delta_n\right\}\\
    \left|S\left(\left\{\Delta_{n}\right\}_{n=0}^{N-1}\right)\right|\le&\prod_{n=0}^{N-1}(1+2\Delta_n) \le \left(1+2\frac{\Delta}{N}\right)^{N}.
\end{align*}
Therefore, we can get the asymptotic upper bound on the number of perturbations as follows:
\begin{align*}
    \left|\left\{s|s\in S(d), d\in D(\Delta)\right\}\right|\le&\left(1+2\frac{\Delta}{N}\right)^{N}\left|D(\Delta)\right|\\
    =&\ O\left(\frac{N^\Delta}{\Delta!}\left(1+\frac{2\Delta}{N}\right)^N\right).
\end{align*}

From the results above, we can compute the perturbation space ratio function $f=T^\Delta/(1+2\Delta/N)^N$ and prove the statement.
\end{proof}
\begin{theorem}
   The perturbation space of temporal encoding is exponentially smaller than that of rate encoding for $T \ge 8$.
\end{theorem}
\begin{proof}
From Lemma~\ref{lemma:pert-space-ratio}, we compute the partial derivative of $\ln f$ over $\alpha$ as follows:
\begin{align*}
    \frac{\partial}{\partial\alpha}\ln f%&= \frac{\partial}{\partial\alpha} N\left[ \alpha T\ln T - \ln (1+2\alpha T) \right]\\
    &=NT\left[ \ln T - \frac{2}{1+2\alpha T} \right].
\end{align*}
Observe that the derivative over $\alpha$ is always positive where $T>e^2\approx7.39$. As $f=1$ at $N=0$ and $T$ is an integer, $f$ increases over $T$ and $N$ monotonically and exponentially for $T\ge8$.
\end{proof}
Note that there still exists $\alpha$, which makes the ratio exponential for $T<8$. Most of our experiments are conducted in $T=5, N=10,$ and $\Delta=1$. In $T=5$ and $N=10$, the space of rate encoding is greater than that of temporal, where $\alpha\gtrapprox 0.05$ and $\Delta=\alpha NT\approx 2.5$.

%\jk{알고리즘 가독성을 위해 line number 를 넣거나 6번째 줄처럼 두 줄로 이어지는 경우를 없애주면 어떨까 합니다.}
%\br{수정하겠습니다!}
\subsection{Direct Counterexample Search}
Using Lemma ~\ref{lemma:pert-space-ratio}, we can predict that verification times of SNNs only depend on $\Delta$ and $N$, which is the number of input neurons. However, the experiment section of our paper shows that verification time increases exponentially as the number of time steps does. Assuming that the verification time inefficiency is from the SMT solver, we propose a direct counterexample search (DCS) algorithm that does not use an SMT solver but brings entirely the same result with an SMT solver. Algorithm~\ref{alg:DCS} describes DCS in pseudocode. This method provides a solution to the exponentially increasing size of the model space of SMT solvers, which grows along with the number of hidden neurons and the number of time steps. This phenomenon is even observed in temporal encoding.
% \sk{두 함수는 분리해서 기술하는 게 나을 것 같습니다. GetPerturb 함수는 재귀 대신 반복문 통해서 기술하는게 더 이해하기 쉬워보입니다. x는 재귀과정에서 사용되지 않는 것 같고요..}
% \br{이부분은 DFS라서 for 문으로만 구현하려면 좀 복잡해 보여서... 수정하지 않고 놔뒀습니다.}
\begin{algorithm}[ht!]
\caption{Direct Counterexample Search}\label{alg:DCS}
\begin{algorithmic}[1]
\Function{GetPerturb}{$\mathbf{x}$, $\Delta$, $i$, $\boldsymbol{\epsilon}$}
    \State $L\gets$ empty array
    \If{$\Delta=0$}
        \State $L$.append$(\mathbf{x}+\boldsymbol{\epsilon})$
    \ElsIf{$i<N$}
        \State $D\gets\{-\min(x_i,\Delta),\ldots,\min(T-1-x_i,\Delta)\}$
        \For{\textbf{each} $\delta\in D$}
            \State $\tilde{\boldsymbol{\epsilon}}\gets\boldsymbol{\epsilon}+\delta\mathbf{e}_i$
            \State $l\gets$ \Call{GetPerturb}{$\mathbf{x},\Delta-|\delta|,i+1,\tilde{\boldsymbol{\epsilon}}$}
            \State $L$.extend($l$)
        \EndFor
    \EndIf
    \State\Return $L$
\EndFunction
\\
\Function{DCS}{$\mathbf{x}$, $y$}
    \For{\textbf{each} $\tilde{\mathbf{x}}\in$ \Call{GetPerturb}{$\mathbf{x},\Delta,0,\mathbf{0}$}}
        \State $\tilde{y} \gets f(\tilde{\mathbf{x}})$
        \If{$y\neq\tilde{y}$}
            \State\Return\textbf{SAT}
        \EndIf
    \EndFor
    \State\Return\textbf{UNSAT}
\EndFunction
\end{algorithmic}
\end{algorithm}

\section{Experiments}
We have implemented the proposed verification algorithm based on the high-performance SMT solver Z3Py~\cite{de2008z3}, the Z3 API in Python developed by Microsoft Research. To train and infer the model, we have used NumPy~\cite{harris2020array}, snnTorch~\cite{eshraghian2021training} and PyTorch~\cite{paszke2019pytorch}, S4NN~\cite{S4NN}. Experiments were conducted on the AMD EPYC 7763 2.45GHz CPU and 1TB of RAM.

To balance the practicality and ease of the experiment, we used the MNIST and FMNIST datasets. Each verification of a model was conducted with 14 fixed inputs and ran once, except for MNIST DCS with $\Delta=1$, which had 5 runs. To get an appropriate performance of the SNNs in reasonable verification time, we set the number of hidden neurons to $20$ in the experiment on time steps and the number of time steps to $5$ on the number of neurons. $\Delta$ was set to 1. The model accuracies of rate encoding range from 52\% to 67\%, and those of temporal encoding range from 70\% to 80\%. The lower model accuracy in rate encoding can affect its mean verification time by the ratio of robust and not robust instances, but we can ignore this because rate encoding was extremely slow even on not robust instances, which is generally faster than robust instances.
\begin{figure}[t!]
\centering
\includegraphics[width=.47\textwidth, clip, trim={0.2cm 0.2cm 0 0.1cm}]{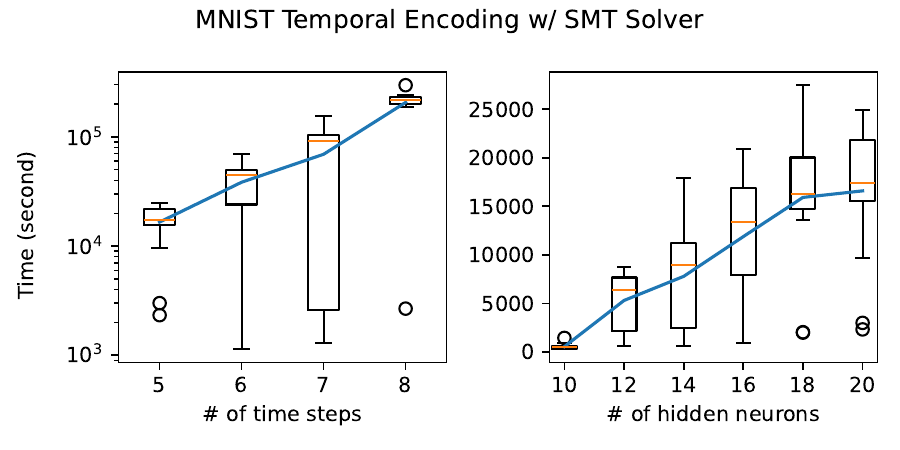}
\includegraphics[width=.47\textwidth, clip, trim={0.2cm 0.2cm 0 0.1cm}]{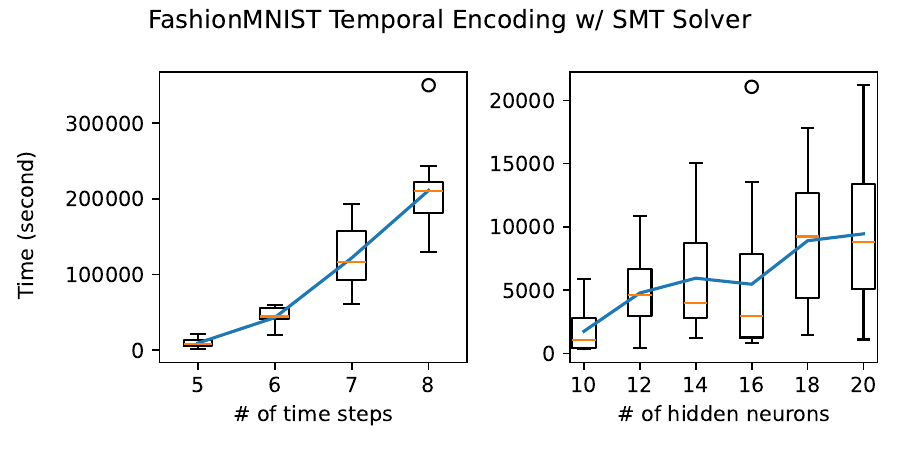}
\caption{Verification runtime with different numbers of time steps (left) and different numbers of hidden neurons (right) at MNIST and FashionMNIST dataset. In FashionMNIST, at 6 time steps, two instances timed out for $>$ 80,000 seconds.
}
\label{fig:verif}
\end{figure}
\paragraph{Verification time with different number of hidden neurons}

With rate encoding, most MNIST instances failed verification in $N\ge12$ within 180000 seconds, and only 8 of 84 instances are verified to be not robust in FashionMNIST within 150,000 seconds. In contrast, Figure~\ref{fig:verif} (right) presents that verification time is very small in temporal encoding.  The verification time does not look like the following exponential time complexity. Using Lemma~\ref{lemma:pert-space-ratio}, we can infer that it is because perturbation space does not depend on the number of hidden neurons.

\paragraph{Verification time with different time steps}

Rate encoding has not had a single successful verification in about 250000 seconds on $T\ge 7$ and has succeeded in only 14 instances in $T\le6$ at the MNIST dataset. Moreover, there was no successful verification in the FashionMNIST dataset in about 150000 seconds. On the contrary, Figure~\ref{fig:verif} (left) shows verification time at temporal encoding over different time steps.  The result differs from the theoretical prediction: temporal encoding performs well even in $T<8$. Verification time in temporal encoding still increases exponentially but shows much less verification time than the baseline. In contrast to lemma~\ref{lemma:pert-space-ratio}, It shows exponential time complexity.

\begin{figure}[t!]
\centering
\includegraphics[width=.47\textwidth, clip, trim={0.2cm 0.2cm 0 0.1cm}]{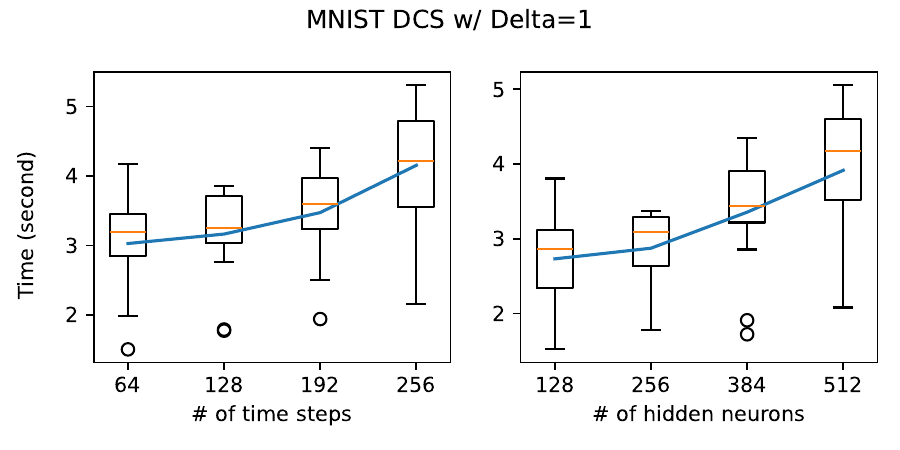}
\includegraphics[width=.47\textwidth, clip, trim={0.2cm 0.2cm 0 0.1cm}]{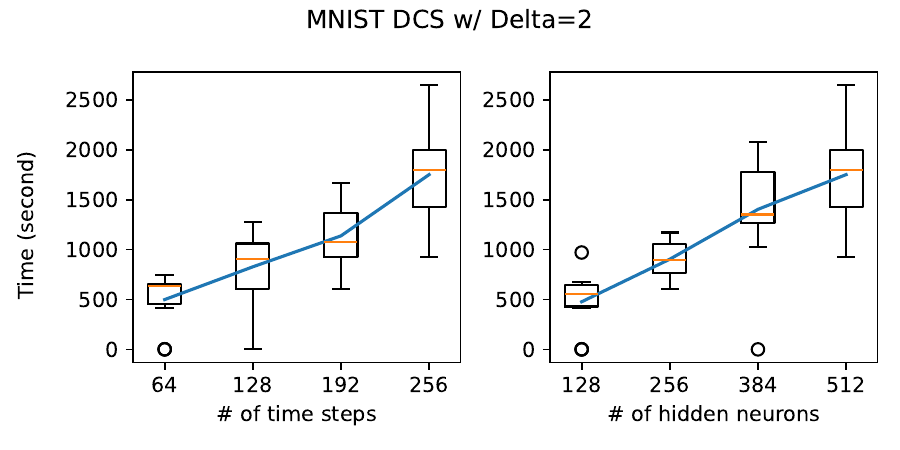}
\caption{DCS Verification runtime with different numbers of time steps (left) and different numbers of hidden neurons (right) at MNIST dataset, with different $\Delta$s.
}
\label{fig:verif-DCS}
\end{figure}
\begin{table*}[t!]
\caption{Experimental results of robustness verification on two models with different numbers of hidden neurons with $\Delta=1$. Values are given as mean $\pm$ standard deviation.}
\centering\label{table:total-model-compare-N}
\begin{tabular}{c|ccccc|ccc}
\hline
Dataset             & \multicolumn{5}{c|}{MNIST}                                                                                                                                        & \multicolumn{3}{c}{FashionMNIST}                            \\ \hline
$|N|$               & \multicolumn{2}{c|}{Instance type}                                                                  & Robust            & Not robust         & Timeout            & Robust            & Not robust        & Timeout             \\ \hline
\multirow{6}{*}{10} & \multirow{2}{*}{Rate}                                                 & \multicolumn{1}{c|}{Result} & 0                 & 14                 & 0                  & 0                 & 1                 & 13                  \\
                    &                                                                       & \multicolumn{1}{c|}{Time}   & -                 & 3657 $\pm$ 2936    & -                  & -                 & 9.7e4             & \textgreater 1.5e5  \\ \cline{2-9} 
                    & \multirow{2}{*}{Tem.}                                                 & \multicolumn{1}{c|}{Result} & 11                & 3                  & 0                  & 3                 & 11                & 0                   \\
                    &                                                                       & \multicolumn{1}{c|}{Time}   & 2660 $\pm$ 1501   & 1616 $\pm$ 337.8   & -                  & 3074 $\pm$ 2350   & 1716 $\pm$ 1377   & \textgreater 3e5    \\ \cline{2-9} 
                    & \multirow{2}{*}{\begin{tabular}[c]{@{}c@{}}Tem.\&\\ DCS\end{tabular}} & \multicolumn{1}{c|}{Result} & 11                & 3                  & 0                  & 3                 & 11                & 0                   \\
                    &                                                                       & \multicolumn{1}{c|}{Time}   & 0.161 $\pm$ 0.002 & 0.089 $\pm$ 0.046  & -                  & 0.184 $\pm$ 0.022 & 0.004 $\pm$ 0.007 & -                   \\ \hline
\multirow{6}{*}{20} & Rate                                                                  & \multicolumn{1}{c|}{Result} & 0                 & 12                 & 2                  & 0                 & 1                 & 13                  \\
                    &                                                                       & \multicolumn{1}{c|}{Time}   & -                 & 4.6e4 $\pm$ 6726   & \textgreater 2.5e5 & -                 & 7.3e4             & \textgreater{}1.5e5 \\ \cline{2-9} 
                    & \multirow{2}{*}{Tem.}                                                 & \multicolumn{1}{c|}{Result} & 12                & 2                  & 0                  & 5                 & 9                 & 0                   \\
                    &                                                                       & \multicolumn{1}{c|}{Time}   & 1.7e4 $\pm$ 6362  & 1.0e4 $\pm$ 7478   & -                  & 1.4e4 $\pm$ 4294  & 1.0e4 $\pm$ 6000  & -                   \\ \cline{2-9} 
                    & \multirow{2}{*}{\begin{tabular}[c]{@{}c@{}}Tem.\&\\ DCS\end{tabular}} & \multicolumn{1}{c|}{Result} & 12                & 2                  & 0                  & 5                 & 9                 & 0                   \\
                    &                                                                       & \multicolumn{1}{c|}{Time}   & 0.168 $\pm$ 0.003 & 0.001 $\pm$ 7.5e-5 & -                  & 0.172 $\pm$ 0.007 & 0.009 $\pm$ 0.015 & -                   \\ \hline
\end{tabular}
\end{table*}
\paragraph{Verification time of DCS}
We found an inconsistency between the theoretical prediction and the experiment results in the time complexity in the number of time steps. We assumed the problem was caused by the inefficiency of the SMT solver, especially by the number of the time step terms. We devised DCS algorithm that directly generates the perturbation set and checks whether any adversarial counterexample exists. Figure~\ref{fig:verif-DCS} shows the verification times of DCS. The number of time steps and hidden neurons are different from the temporal encoding experiment. This is because the verification times were too low, so the results could be noisy.

We also conducted a simple experiment in DCS, using the perturbation space as inputs and the original prediction as labels. We succeeded in making not robust instances to be robust.

\paragraph{Comparison of verification methods}
Table~\ref{table:total-model-compare-N} compares verification methods with rate SMT, temporal SMT, and DCS with temporal encoding in two datasets. These verifications were done with the model with 10 or 20 hidden neurons and $T=5$, and $\Delta=1$. The accuracy of the rate model and temporal model was 52\% and 59\% in 10 hidden neurons and 61\% and 79\% in 20 hidden neurons, respectively. Despite the tendency that a robust sample requires a longer time~\cite{BanerjeeGBM23} for verification, there is a huge gap between temporal encoding and rate encoding, and the gap gets much bigger when compared with the proposed DCS algorithm.

% Table~\ref{table:total-model-compare-delta} illustrates the verification times of three methods across a range of $\Delta$ values. The verification times demonstrate an exponential increase across all methods. In $\Delta=4$, no robust instance was identified, even in the context of our DCS approach. Considering the perturbation space of temporal encoding from lemma~\ref{lemma:pert-space-ratio}, the robust instance verification time in DCS is estimated to be approximately 2.4e7 seconds. However, empirical verification times have been observed to increase approximately 300-fold for each incremental increase in $\Delta$. Consequently, it also can be predicted that these times would be around 5.1e6 seconds.

\section{Conclusions and Future Work}
We have proposed the SMT encoding of the temporally encoded SNNs for efficient robustness verification of SNNs. The key contributions of the paper are to exponentially reduce the size of perturbation space by exploiting temporal encoding of input instead of rate encoding and to reduce the inefficiency of the SMT solver. We have achieved the exponential reduction in the perturbation space by restricting each neuron to fire only once, which we have proved theoretically while also providing empirical evidence. However, as the time step increases, we observe an exponential increase in verification time, which is contrary to our prediction and can be interpreted as the SMT solver's inability to narrow the search space properly. So we designed the DCS algorithm and succeeded in making the verification times barely depend on the number of time steps. We also succeeded in getting reasonable verification time in large models, and now we can also use GPUs to accelerate formal verification of SNNs due to separable SNN inference.

While this study has achieved impressive progress in the formal verification of SNNs, further research is still needed. Despite the verification time dependency on the number of input neurons, we could not try formal verification to more input neurons because of the training difficulty of temporal SNNs. Additionally, experiments on more complex datasets and more complex network architectures, such as convolutional SNNs, are necessary. We tried to train SNNs in cifar10~\cite{cifar}, but without convolutional architecture, it was hard to train. In view of adversarial training, we also have challenges. Fine-tuning the network with the counterexamples will help the adversarial training of SNNs.

\bibliography{aaai25}

\newpage

\onecolumn
\appendix

\section{Appendix}

Here we provide detailed information about the experiments introduced in the main paper, which could not be included in the main text due to space constraints.

\subsection{Hyperparameter Setting for Experiments}
\begin{table}[h]
\centering
\begin{tabular}{l|c}
\hline
{\bf Hyperparameter Name}                             & {\bf Value} \\ \hline
Learning rate of baseline SNN & 5e-4  \\ \hline
Learning rate of temporal SNN & 0.2   \\ \hline
Target firing time ($\gamma$) of temporal SNN         & 3     \\ \hline
Initial weight range of the first layer of temporal SNN  & [0, 5]    \\ \hline
Initial weight range of the second layer of temporal SNN & [0, 50]    \\ \hline
\end{tabular}
\caption{Hyperparameter settings used in experiments.}
\label{tab:hyperparameter}
\end{table}

Table~\ref{tab:hyperparameter} provides the hyperparameter settings used in our experiments. We use the Adam optimizer~\cite{KingmaB14} with a learning rate of 5e-4 for training the baseline SNN.
Following the configurations from S4NN~\cite{S4NN}, we set the learning rate of temporal SNN to 0.2 and initialize the input-hidden
and hidden-output synaptic weights with random values drawn from uniform distributions in the ranges $[0, 5]$ and $[0, 50]$. Note that we use the off-the-shelf optimizer for S4NN, which is publicly available\footnote{https://github.com/SRKH/S4NN}, for training the temporal SNN in our experiments.

\subsection{DCS Experiment Results}

In this section, we report the experimental results of the proposed DCS algorithm in more detail.

\begin{table}[h]
\centering
\begin{tabular}{cc|cc|cc}
\hline
\multicolumn{2}{c|}{\bf \# of time steps}
& \multicolumn{2}{c|}{64}             & \multicolumn{2}{c}{128}        \\ \hline
\multicolumn{2}{c|}{\bf Instance type}                        & Robust            & Not robust      & Robust            & Not robust \\ \hline
\multicolumn{1}{c|}{\multirow{2}{*}{\bf $\Delta=1$}} & Result & 11                & 3               & 13                & 1          \\
\multicolumn{1}{c|}{}                            & Time   & $1.968\pm 0.605$   & $0.885\pm 0.811$ & $3.151 \pm 0.939$ & $0.002 \pm 0.000$      \\ \hline
\multicolumn{1}{c|}{\multirow{2}{*}{$\Delta=2$}} & Result & 11                & 3               & 13                & 1          \\
\multicolumn{1}{c|}{}                            & Time   & $635.2 \pm 85.17$   & $0.799 \pm 0.581$ & $891.4 \pm 281.8$ & $0.002$      \\ \hline\hline
\multicolumn{2}{c|}{\bf \# of time steps}                     & \multicolumn{2}{c|}{192}             & \multicolumn{2}{c}{256}        \\ \hline
\multicolumn{1}{c|}{\multirow{2}{*}{$\Delta=1$}} & Result & 14                & 0               & 14                & 0          \\
\multicolumn{1}{c|}{}                            & Time   & $4.004 \pm 1.175$ & -               & $5.145 \pm 1.910$ & -          \\ \hline
\multicolumn{1}{c|}{\multirow{2}{*}{$\Delta=2$}} & Result & 14                & 0               & 14                & 0          \\
\multicolumn{1}{c|}{}                            & Time   & 1138 $\pm$ 324.9  & -               & $1752 \pm 422.5$  & -          \\ \hline
\end{tabular}
\caption{Verification results of SNNs with 512 hidden neurons with various numbers of time steps.}
\label{tab:various_ts}
\end{table}

Table~\ref{tab:various_ts} shows the experimental results of SNNs with a fixed number (512) of hidden neurons on the MNIST benchmark. We only vary the number of time steps for spike trains to see how the verification time changes depending on the number of time steps. The values are presented in the table as the mean $\pm$ standard deviation of the validation times for a total of 14 samples. Verification times increase linearly over the number of time steps at $\Delta=2$. On the other hand, the linear relationship between the number of time steps and verification time is not clearly observed in the case of $\Delta=1$. We speculate the linear relationship is not well observed because of the signal-to-noise ratio in the case of $\Delta = 1$ due to the small absolute magnitude of the values.

\begin{table}[h]
\centering
\begin{tabular}{cc|cc|cc}
\hline
\multicolumn{2}{c|}{\bf \# of hidden neurons}               & \multicolumn{2}{c|}{128}              & \multicolumn{2}{c}{256}        \\ \hline
\multicolumn{2}{c|}{\bf Instance type}                        & Robust            & Not robust        & Robust            & Not robust \\ \hline
\multicolumn{1}{c|}{\multirow{2}{*}{$\Delta=1$}} & Result & 11                & 3                 & 14                & 0          \\
\multicolumn{1}{c|}{}                            & Time   & $1.355 \pm 0.211$ & $0.628 \pm 0.864$ & $2.184 \pm 0.393$ & -          \\ \hline
\multicolumn{1}{c|}{\multirow{2}{*}{$\Delta=2$}} & Result & 11                & 3                 & 13                & 1          \\
\multicolumn{1}{c|}{}                            & Time   & 607.4 $\pm$ 138.2 & $0.484 \pm 0.654$ & $927.2 \pm 143.4$ & $609.2$      \\ \hline\hline
\multicolumn{2}{c|}{\bf \# of hidden neurons}                 & \multicolumn{2}{c|}{384}              & \multicolumn{2}{c}{512}        \\ \hline
\multicolumn{1}{c|}{\multirow{2}{*}{$\Delta=1$}} & Result & 13                & 1                 & 14                & 0          \\
\multicolumn{1}{c|}{}                            & Time   & $4.678 \pm 1.384$ & $0.010 \pm 0.011$             & $5.145 \pm 1.910$ & -          \\ \hline
\multicolumn{1}{c|}{\multirow{2}{*}{$\Delta=2$}} & Result & 13                & 1                 & 14                & 0          \\
\multicolumn{1}{c|}{}                            & Time   & $1512 \pm 320.6$  & $0.013$             &$1752 \pm 422.5$   & -          \\ \hline
\end{tabular}
\caption{Verification results of SNNs with 256 time steps with various numbers of hidden neurons.}
\label{tab:various_hn}
\end{table}

Table~\ref{tab:various_hn} is the experimental results of SNNs with a fixed number of time steps (256) but with different numbers of hidden neurons. We can also observe the linear relationship between the number of hidden neurons and verification runtime as in Table~\ref{tab:various_ts}. 
% Verification times are unstable, but it does not increase exponentially over the number of hidden neurons in each $\Delta$ case.

\end{document}